\documentclass{article}

\usepackage{PRIMEarxiv}
\usepackage[utf8]{inputenc} % allow utf-8 input
\usepackage[T1]{fontenc}    % use 8-bit T1 fonts
\usepackage{url}            % simple URL typesetting
\usepackage{booktabs}       % professional-quality tables
\usepackage{amsfonts}       % blackboard math symbols
\usepackage{nicefrac}       % compact symbols for 1/2, etc.
\usepackage{microtype}      % microtypography
\usepackage{lipsum}
\usepackage{fancyhdr}       % header
\usepackage{graphicx}       % graphics
\graphicspath{{media/}}     % organize your images and other figures under media/ folder
\usepackage{changepage} % put in preamble
\usepackage{tabularx} % add this to your preamble
\usepackage{multirow} % in preamble
\usepackage{array}    % for >{\raggedright} alignment
\usepackage{graphicx}
\usepackage{pgf-pie}  % Add this to your preamble
\usepackage{float}
\usetikzlibrary{trees}
\usepackage{amsmath}
\usepackage{graphicx}      % graphics
\usepackage{amsmath}       % math
\usepackage{booktabs}      % tables
\usepackage{hyperref}      % hyperlinks (load last!)

\hypersetup{
    colorlinks=false,      % disables colored links completely
    pdfborder={0 0 0},     % removes boxes around links
}
\title{Simulating Meaning, Nevermore! \\ Introducing ICR: A Semiotic-Hermeneutic Metric for Evaluating Meaning in LLM Text Summaries}
\author{
  Natalie Perez \\
  University of Hawai‘i \\
  Hawai‘i, USA \\
  \texttt{natalie.perez@hawaii.edu} \\
  \And
  Sreyoshi Bhaduri \\
  Private Corporation \\
  New York, USA \\
  \texttt{sreyoshibhaduri@gmail.com}
    \And
   Aman Chadha\thanks{This research was conducted independently of the author’s work at Amazon.}  \\
  Amazon GenAI \\
  California, USA \\
  \texttt{hi@aman.ai}
}
\begin{document}
\maketitle

\begin{abstract}
Meaning in human language is relational and context-dependent, and it emerges through a dynamic system of signs rather than fixed interactions between words and concepts. In computational computing, the inherently sembiotic and interpretative structures in language make generating accurate meaning challenging. This article advances an interdisciplinary framework for studying meaning in machine-generated language by integrating semiotics and hermeneutics via systematic qualitative research methods. We first review scholarship on meaning and machines, tracing how linguistic signs are transformed into vectorized representations in both static and contextualized embedding models, and identifying gaps between statistical approximation and human interpretive meaning. Building on this foundation, we introduce the Inductive Conceptual Rating (ICR) metric, a qualitative evaluation approach grounded in inductive content analysis and reflective thematic analysis, designed to assess semantic accuracy and meaning alignment in generative artificial intelligence (GenAI) outputs, beyond surface-level lexical and similarity metrics. We then apply ICR in an empirical study, comparing large language model (LLM) generated thematic summaries with human-generated outputs across five datasets (N = 50 to 800). The results show that while language models achieve high linguistic and semantic similarity scores, they consistently underperformed when compared with human outputs on semantic accuracy, particularly in capturing recurring, contextually grounded meanings. However, LLM performance improves with larger datasets, but the results remains variable across models, and we hypothesize the variability in model results is associated with the increase in frequency of reoccurring or related concepts. We conclude by discussing the epistemological and practical implications of treating GenAI outputs as dynamic sign systems, arguing for evaluation frameworks that foreground human interpretation when assessing meaning with respect to machine-mediated language generation. \\
\end{abstract}

% keywords can be removed

\section{Introduction}

\begin{center}
``On the morrow he will leave me, as my Hopes have flown before.'' \\
Then the bird said ``Nevermore.'' [...] \\
``Take thy beak from out my heart, and take thy form from off my door!'' \\
Quoth the Raven ``Nevermore.'' \\
``And my soul from out that shadow that lies floating on the floor \\
Shall be lifted, nevermore!'' \\
\textit{--- The Raven by Edgar Allan Poe, 1845}
\end{center}

\vspace{2em} % <-- adds vertical space before the introduction

Language meaning is dependent on signs, interrelations, and context. Saussure (1916) described this relationship as a linguistic sign system {\cite{DeSaussure1916}; this system always contains the combination of a signifier (word) and a signified (conceptual meaning), with its sign (overall meaning) generated based on how the signifier and signified relate within a system of signs. This argument is easily illustrated within the poem \textit{The Raven}, written by Edgar Allan Poe in 1845 {\cite{Poe2013}. The word ``\textit{Nevermore}'' is a signifier (word), yet the word is repeated throughout \textit{The Raven}, and its signified (conceptual meaning) changes with each new stanza. In one section of the poem, ``\textit{Nevermore}'' refers to a lover never getting to see a lost love again, and a few stanzas later, ``\textit{Nevermore}'' is used to refer to a person struggling to find comfort, and then it later is used to refer to a person experiencing eternal despair. Within the poem, \textit{Nevermore}’s meaning changes, rather than staying the same, making \textit{Nevermore} an example of a polysemy, which is one signifier (word) linked to multiple signifieds (meaning), within a single text \cite{PurbaDamanik2025}. In contrast, large language models (LLMs) are often evaluated using automated evaluation metrics, which often treat signifiers (words) as static, context-independent units, which assumes that a word carries a fixed signified (conceptual meaning) \cite{Bhargavi2025}. As a result, LLM evaluation metrics might not always capture the context-dependent shifts in signifieds, producing outcomes that risk failing to capture the nuanced, relational and fluid meaning of signs present in texts like \textit{The Raven}. It is along these lines that we argue the use of automated metrics as computational proxies for meaning could result in metric scores that do not always reflect the intended meaning of language within a reference text. \\

Context-dependent meaning in human language directly illuminates a central challenge in computational linguistics and natural language processing (NLP). For instance, in NLP, a common strategy to model language meaning has been word embeddings that encode signifiers (words) as vectors (points) in a high‑dimensional semantic space \cite{Apidianaki2023}. These embeddings approximate semantic relationships, which are words that frequently appear in similar contexts; an embedding places the related words or tokens closer together in a vector space, which allows models to “predict” likely words or phrases in context \cite{Izzidien2022, IzzidienStillwell2021}. However, unlike human interpretation, traditional word embeddings do not inherently encode the signified (meaning). In traditional static embeddings (e.g., BoW, TF-IDF, Word2Vec, GloVe), models assign a single fixed vector to each word, regardless of context \cite{Bhargavi2025}. Unfortunately, while these traditional models can create dense word vectors that relate to general semantic relationships, they fail to distinguish across the different signifiers (meanings), and can face particular challenges with polysemous words across different contexts \cite{Bhargavi2025, PurbaDamanik2025, Grindrod2024}. \\

More recent advances have attempted to overcome the limitations of static embeddings by developing contextualized embedding (e.g., BERT, RoBERTa, DistillBERT, ELECTRA, ALBERT) that generate context‑dependent or dynamic word representations. Contextual embeddings are designed to identify the meaning of a word based on the surrounding words, which has helped to address polysemy words and richer syntactic and semantic relationships \cite{Bhargavi2025}. Not surprisingly, studies applying contextual embeddings have reported improvements in topic modeling \cite{Viegas2025, Alizadeh2025}, text summarization \cite{GangundiSridhar2025}, sentiment analysis \cite{PaneruThapa2025}, etc. Despite these improvements, even contextual embeddings remain approximations, as illustrated by automated metrics, as the embedded signifiers (words) and signified (meaning) are based on statistical patterns and co‑occurrence frequencies rather than grounded, embodied human experience or the cultural, historical, or social nuances that constantly shape a word’s meaning \cite{ArsenievKoehler2024, DeSaussure1916}. Consequently, while these new embedding strategies provide powerful and new computational tools for language generation, these models continue to lack the ability to fully replicate the depth, variability, and contextual richness of human interpretive processes. \\

\section{Background}
Researchers frequently evaluate GenAI outputs using automated metrics designed to quantify textual characteristics such as similarity, coherence, or alignment. Table 1 summarizes prominent metrics across a range of output categories (Table 1), as illustrated by Croxford et al. (2025) \cite{Croxford2025}. \\

\begin{table}[ht!]
\centering
\caption{List of Common LLM Evaluation Metrics, Applications, and Limitations (Adapted from Croxford et al., 2025)}
\label{tab:llm_metrics}
\small
\begin{tabularx}{\textwidth}{>{\raggedright\arraybackslash}p{2.5cm} 
                              |>{\raggedright\arraybackslash}p{2.5cm} 
                              |>{\raggedright\arraybackslash}X 
                              |>{\raggedright\arraybackslash}X 
                              |>{\raggedright\arraybackslash}X}
% \hline
\toprule
\textbf{Reference Details} & \textbf{Type} & \textbf{Metrics} & \textbf{Application} & \textbf{Metric Limitations} \\ 
\toprule
% \hline

\multirow{4}{*}{\parbox{2.5cm}{Reference Dependent}} & Word or Character Based & ROUGE, METEOR, JS Divergence, CIDEr, PyrEval, sacreBLEU, SERA, POURPRE, BE, BLEU, GTM, WER/TER, ITER/CDER, chF, characTER, EED, YiSi, Q-Metrics & Evaluate lexical/character overlap and structural fidelity. & Does not measure context-dependent meaning, polysemy, or paraphrases; limited semantic nuance. \\ \cmidrule{2-5}

 & Learned Metrics & COMET, BLEURT, CREMER, BEER, BLEND, Composite, NNEval, RUSE, BERT & Evaluate model semantic adequacy and fluency via learned representations. & Does not measure implied meaning, rhetoric, or cultural/contextual cues. \\ \cmidrule{2-5}

 & Probability-Based & CBMI, NIST & Weigh matches by informativeness, emphasizing rare/content-rich n-grams. & Does not measure factual correctness or real-world grounding. \\ \cmidrule{2-5}

 & Embedding-Based & BERTScore, MoverScore, AutoSumm ENG, MeMoG, SPICE, BERTr, WEWPI, WMD, SIMILE, NUBIA & Evaluate semantic similarity via embeddings. & Does not measure pragmatic, interpretive, and ethical dimensions. \\ \hline

\multirow{4}{*}{\parbox{2.5cm}{Reference Independent}} & Word or Character Based & SIMetrix & Assesses intrinsic structural and stylistic properties. & Does not measure meaning, factuality, or task-specific adequacy. \\ \cmidrule{2-5}

 & Learned Metrics & COMET & Predicts fluency, coherence, and adequacy without references. & Does not measure nuanced meaning or ethical alignment. \\ \cmidrule{2-5}

 & Probability-Based & BARTScore, HARiM+ & Evaluates likelihood under pretrained LMs for fluency/coherence. & Does not measure factual accuracy or contextual correctness. \\ \cmidrule{2-5}

 & Embedding-Based & SEM-nCG & Evaluates semantic organization, discourse coherence, and content structure. & Does not measure model reasoning, argumentation, or multi-document coherence. \\ 
 % \hline
 \bottomrule

\end{tabularx}
\end{table}

While these metrics have been used to scale, reproduce, and examine a variety of linguistic characteristics, they have limitations. For instance, automated metrics tend to identify surface-level language or statistical proximity over interpretive meaning, which can fail to account for polysemy, contextual shifts, rhetorical intent, or culturally situated interpretations \cite{Karjus2025, Bender2020}. These outcomes can be problematic, since GenAI outputs may achieve high automated scores, even with subtle but substantive distortions in the actual meaning, emphasis, or truthfulness of a reference text \cite{MillerTang2025}. We argue that the use of automated metrics as computational proxies for meaning can result in scores that do not accurately or truthfully reflect input data meaning.\\

\subsection{Traditional Qualitative Evaluation Approaches}

In contrast, traditional qualitative evaluation approaches include a range of approaches to evaluate meaning. These approaches are embedded in specific research designs, such as reflexive thematic analysis, inductive content analysis, phenomenological analysis, discourse analysis, narrative analysis, semiotic analysis, and hermeneutic inquiry, among others, to evaluate meaning as emergent, relational, and context-dependent, emphasizing how individuals use language to construct experiences, identities, power, and understanding \cite{BraunClarke2006, SmithFlowersLarkin2009, Fairclough1995, Riessman2008, Eco1979, Gadamer1975}. Additional approaches include expert review panels, rubric-based coding schemes, and consensus scoring, which often utilize predefined categories, a priori criteria, or standardized codebooks to support consistency and comparability \cite{MilesHubermanSaldaña2014, Guest2011}. While these approaches offer rich sensitivity to context, and are arguably better suited to identify meaning characteristics, like polysemy, cultural nuances, and idiographic positionalities, they are typically descriptive, case-bound, and are not designed to explicitly replicate reproducible metrics \cite{MilesHubermanSaldaña2014, BraunClarke2021}. As a result, traditional qualitative evaluation excels at interpreting meaning, but not necessarily scaling insights \cite{Guest2011, CreswellPoth2016}. \\

\subsection{Human-in-the-loop (HITL) Approaches}

Human-in-the-loop (HITL) approaches often complement automated evaluation metrics, but are arguably less rigorous than traditional qualitative methods in explicitly addressing how meaning is constructed, negotiated, or interpreted across contexts. The literature illustrates the ways HITL contributes to evaluating language models through active learning, interactive machine learning, and expert-guided feedback by supplying labels, correcting predictions, and shaping model learning \cite{WagnerKueblerZalnieriute2025}.  HITL approaches are also used to support real-time intervention when AI outputs are ambiguous or potentially biased \cite{Mamman2024}. Aside from providing judgment, in many cases, HITL approaches focus on annotations, label creation or confirmation, or preference rankers, guiding model training and validation without an explicit account of how meaning is constructed or negotiated across contexts \cite{Amershi2014}. In addition, risk-based frameworks use humans to calibrate involvement according to task criticality and potential harm \cite{KandikatlaRadeljic2025}. Overall, HITL integrates human judgment to improve performance, fairness, and safety, but typically does not operationalize interpretive meaning explicitly. \\

\subsection{Proposal of a New Epistemological Evaluation Metric}

Consequently, there is a need to adequately identify how word meanings vary across contexts to more comprehensively evaluate the truthfulness of GenAI outputs, particularly from an epistemological perspective on meaning. This study introduces a methodology for systematic qualitative research that leverages a two step analytical process using a Reflective Thematic Analysis (RTA) and an Inductive Content Analysis (ICA) to generate a metric for quantifying the semantic fidelity of a GenAI output when compared with a reference text. An Inductive Conceptual Rating (ICR) provides a structured, rigorous approach for identifying signifiers (words), signifieds (conceptual meaning), categorizing emergent themes, and interpreting language within context, without imposing predefined relationships, calculations, codes, or embedded assumptions \cite{EloKyangas2008, Mayring2014}. Grounding this approach within the context of semiotics and hermeneutics allows this approach to emphasize relational and context-dependent, highlighting the centrality of human interpretation in understanding how meaning is produced, interpreted, and negotiated in language (see Appendix 1 for more conceptual details). \\

This study contributes theoretically by framing meaning in language as relational, context-dependent, and emergent, emphasizing the limitations of current computational metrics and LLM embeddings in capturing nuanced, culturally situated meaning \cite{DeSaussure1916, Eco1979, Gadamer1975, ArsenievKoehler2024}. Methodologically, it introduces the Inductive Conceptual Rating (ICR), combining reflective thematic and inductive content analysis to systematically evaluate meaning fidelity and truthfulness in LLM-generated outputs without relying on predefined codes \cite{EloKyangas2008, Mayring2014}. Empirically, applying ICR across five datasets (N = 50 to 800) reveals that LLMs, while achieving high lexical or embedding-based similarity, consistently underperformed, when compared with humans, in capturing contextually grounded and recurring meanings. While LLM performance showed improvements with larger datasets, the results varied. \\

\section{Introducing an Interpretative Evaluation Metric: Inductive Conceptual Rating (ICR)}
	
Evaluating meaning in GenAI outputs is challenging. For instance, automated metrics, like lexical overlap, embedding similarity, or probability-based scores provide scale, reproducibility, and top-down, deductive evaluation, but they often fail to capture context-dependent meaning, polysemy, or emergent semantics. Human-in-the-loop (HITL) approaches offer functional judgment, but they typically lack fully systematic, epistemologically grounded evaluation. In contrast, traditional qualitative methods provide interpretive depth and relational understanding, though they are less scalable \cite{BraunClarke2021}. In the context of LLM evaluation, automated metrics operationalize a top-down strategy, flagging misalignments as errors based on predefined patterns, whereas human-centered, case-based approaches, adopt a bottom-up, inductive strategy, allowing researchers to uncover the nuanced, context-dependent ways meaning is constructed, changed, or omitted. \\

\subsection{Epistemological Justification}

According to Borgstede and Scholz (2021), differences in quantitative and qualitative approaches mirror variable-based versus case-based modeling \cite{Borgstede2021}. In variable-based quantitative modeling, relational structures are represented as functional relations across all objects \cite{Borgstede2021}:

{\Large
\[
\forall i: \quad y_i = f(x_i)
\]
}
\\
This equation expresses that for every instance \(i\) in the dataset, \(y\) is a function of \(x\). Evaluation takes place deductively, where predictions for individual instances are derived from general law, and deviations indicate potential flaws in the underlying theory or model \cite{Borgstede2021}. In this context, ablation of features can take place, like removing n-grams, embeddings, or likelihood estimates, to produce predictable changes in scores. In the context of LLM evaluation metrics, this reflects the structural sensitivity that a metric might have, rather than subtleties in its ability to differentiate meaning \cite{ZhangPeng2024}. When replications fail or results vary, it does not necessarily indicate semantic instability, but rather that the assumed one-to-one mapping between signifier and signified does not hold to be universally true \cite{Borgstede2021}. In other words, if an evaluation metric gives inconsistent results or fails to replicate, this variability could show that a respective metric's reliability or general applicability is limited.\\

Conceptually, this deductive framing treats signs as stable inputs rather than as as interpretive or context-dependent relationships. For instance, a signifier (e.g., tokens, n-grams, embeddings, or likelihood estimates) is treated as a fixed, measurable input \(x\), while the signified is assumed to follow deterministically as an output \(y = f(x)\). As a result, the sign itself is reduced to a functional mapping between linguistic form and evaluative outcome, rather than understood as a dynamic relation shaped by context, interpretation, or use. The output, therefore, is that meaning is modeled not as a dynamic, relational process but as a predictable outcome of feature extraction, reducing semiotic complexity to functional determinism and constraining how language, judgment, and understanding can be represented. \\

Alternatively, in case-based qualitative modeling, relational structures are expressed as existential statements \cite{Borgstede2021}:

{\Large
\[
\exists i: XYZ_i
\]
}
\\
This equation expresses that there exists at least one case \(i\) for which a given property or pattern holds true. Unlike universal claims, existential statements like this one do not deductively generalize with other cases; instead, any extension beyond the observed instance is inductive, relying on perceived similarities rather than invariant laws \cite{Borgstede2021}. This means that the variance of $XYZ$ across cases is unconstrained and its distribution may be sparse, multimodal, or even idiosyncratic. Any extension of the finding beyond the observed case, therefore, takes place inductively rather than deductively. In qualitative modeling, ablation may occur through coder variation, shifts in conceptual granularity, or differences in thematic overlap, each of which can yield divergent yet interpretable outcomes \cite{Bedemariam2025Potential, BraunClarke2022}; these differences are not noise relative to the claim but are instead consistent with the logic of the equation, which does not assume stability or repeatability. From this lens, variability would only constitute a problem if a universal claim were being made; under an existential formula, however, variation is epistemologically meaningful, as it can help to identify boundary conditions, reveal heterogeneity, and clarify where $XYZ$ does not occur. In this sense, $\exists i : XYZ_i$ defines a logical space in which divergence across cases is not only expected but can also be analytically productive, particularly within qualitative modeling. Hence, rather than signaling error, variability, within this approach, is epistemologically informative, as it reveals the boundaries of semantic generalization, clarifies the scope of intended application, and identifies domains where top-down metrics may fail to capture critical nuances \cite{Borgstede2021, CreswellPoth2016, Creswell2017, BraunClarke2006}. \\

Conceptually, this inductive framing treats signs, signifiers, and signifieds as relational and contingent rather than as fixed one-to-one mappings. For instance, a signifier is not assumed to carry a single, stable meaning across cases; instead, its interpretation may shift depending on culture, context, conceptual granularity, or analytical lens \cite{CreswellPoth2016, SmithFlowersLarkin2009}. Consequently, the signified is context-dependent and emergent, formed through interpretation rather than deduced from a universal rule \cite{Creswell2017}. The sign itself is, therefore, understood as a situated relationship between signifier and signified, co-constructed within a specific case rather than mechanically reproducible across all cases \cite{DeSaussure1916}. \\

The core takeaways, regarding the differences in epistemological roles of quantitative and qualitative approaches, is that quantitative methods aim to demonstrate structure, reproducibility, and generalizable claims, and also indicate whether a model or metric behaves consistently across datasets; whereas, qualitative methods aim to capture nuance, context, and interpretative variability, showing where meaning is fluid and dependent on a range of possible factors. Overall, these approaches are complementary, particularly for LLM evaluation, as quantitative metrics are effective at measuring lexical or surface-level features, such as word frequencies, embeddings, or n-grams, and at testing patterns reproducibly across datasets. However, qualitative evaluation excel at revealing semantic meaning, nuance, and contextual variation, showing how meaning is situated, tolerating variability and highlighting areas where quantitative metrics may overlook subtle or context-dependent aspects of meaning. By combining both approaches, researchers can leverage the precision and reproducibility of quantitative metrics while also accounting for the richness and interpretive flexibility revealed through qualitative analysis. Quantitative methods identify what patterns exist, and qualitative methods clarify what those patterns mean and where they might fail, providing a more holistic evaluation of meaning in LLM outputs.

\subsection{ICR Metric}

Given the context-dependent nature of both human language and model-generated text, evaluating meaning in GenAI outputs requires systematic, interpretive methods that are grounded in accurate representations of meaning. The Inductive Conceptual Rating (ICR) metric is designed to address automated evaluation metric gaps by combining two traditional qualitative approaches: Reflective Thematic Analysis (RTA) \cite{BraunClarke2021} and Inductive Content Analysis (ICA) \cite{ZhangWildemuth2009}. By merging these distinct qualitative approaches, researchers can utilize their respective outputs and quantify a transparent and theoretically grounded evaluation of meaning. More specifically, the ICR metric leverages the interpretive role of human experts by situating meaning within a relational and context-dependent framework that quantifies how closely GenAI outputs identity and illustrates meaning; ICR does so by explicitly modeling signifiers (words) and signifieds (meanings), using thematic structures, to capture semantic nuances and tracing those structures across outputs. \\

\subsection{Analytical Overview}

This approach integrates Braun and Clarke’s (2021) Reflective Thematic Analysis (RTA) \cite{BraunClarke2021} with Zhang and Wildemuth’s (2009) Inductive Content Analysis (ICA) \cite{ZhangWildemuth2009},  to establish a grounded interpretive baseline in which GenAI outputs can be evaluated for accuracy and truthfulness. The the following procedures outline how to deploy the ICR metric to evaluate content generated from LLMs, using reference text(s): \\

\subsubsection{Step 1. Conduct a Reflective Thematic Analysis (RTA)}

The first stage of ICR involves conducting a Reflective Thematic Analysis (RTA) across the entire reference dataset, or “golden datasets,” under examination. RTA is used to inductively identify patterns of meaning, recurrent themes, and contextual relationships through six steps: data familiarization, generation of initial codes, identification of themes, thematic review, further defining themes, and generating insights \cite{BraunClarke2022}. This process emphasizes extensive engagement with the raw, unstructured data, engaging in researcher reflexivity, interpretive depth, and achieving coding and meaning saturation \cite{HenninkKaiser2022, BraunClarke2022}. \\

To enhance reliability and rigor, multiple expert researchers independently analyze the dataset, comparing interpretations and resolving discrepancies through discussion or statistical analysis. In particular, statistical measures such as Cohen’s kappa \cite{cohen1960kappa}, Krippendorf's alpha \cite{krippendorff2018content} or Fleiss’ kappa \cite{fleiss1971measuring} can be applied to quantify inter-rater agreement, providing an empirical check on the consistency of thematic coding and alignment. This systematic and collaborative approach produces a human-interpreted thematic structure that reflects how meaning is constructed, negotiated, and stabilized within the dataset. Importantly, RTA prioritizes conceptual coherence, contextual nuance, and interpretive insight over frequency-based dominance. The resulting themes function as an interpretive baseline of words, meanings, conceptual sequencing, and relationships grounded in human sense-making, that will later be used to compare and evaluate GenAI outputs. \\

\subsubsection{Step 2. Conduct an Inductive Content Analysis (ICA)}

Once the RTA baseline has been established, the second stage of ICR applies an Inductive Content Analysis (ICA) to examine GenAI outputs. Unlike RTA, which produces a human-interpreted thematic structure, ICA is inductive and exploratory, allowing the GenAI outputs to reveal how meanings are represented \cite{Mayring2014}. Researchers systematically examine outputs to identify signifiers, emergent categories, and patterns of meaning without imposing the RTA-derived themes a priori. In particular, this process involves documenting and analyzing the concepts generated by GenAI outputs. This process ensures that model-generated interpretations are captured on their own terms, reflecting how a model constructs semantic relationships, prioritizes certain concepts over others, and makes decisions about reference text nuances like polysemy words. \\

Importantly, an ICA must be independently conducted and applied to both the researcher-generated RTA output and the GenAI output. To perform an ICA, scholars can use Zhang and Wildemuth’s (2009) qualitative content analysis framework \cite{ZhangWildemuth2009}, which includes eight-steps: preparing data, defining unit of analysis, developing categories/coding scheme, testing coding scheme, coding full dataset, assess consistency, analyzing and drawing conclusions, and producing findings. Importantly, the same systematic process must be applied to both the human-generated output (RTA) and model-generated output (LLM) (refer to Appendix 2 for example output). \\

\subsubsection{Step 3. Compare the GenAI and RTA Outputs via ICA}

Once the ICA is individually applied to the human-generated and model-generated outputs, researchers can examine the presence or absence of previously identified categories or concepts to analytically compare the GenAI outputs with the RTA-derived human baseline. This comparison enables researchers to evaluate the alignment, distortion, omission, or fabrication of meaning in relation to a reference text(s). \\

For example, consider an RTA output that identifies the theme “Work-life Harmony Challenges”, with associated concepts like “Flexible Scheduling,” “Personal Well-being,” and “Family Support.” A systematic comparison using the ICA process might reveal that the GenAI output overemphasizes career-related aspects while under-representing personal well-being and family support. In another dataset with the same theme, the comparison might show that the GenAI output generates novel conceptual combinations not present in the human data, producing a hallucination error. Alternatively, in a different dataset, the GenAI output may closely match the human-RTA baseline. Without a systematic baseline for comparison, it is difficult to determine how accurate or truthful a model’s outputs are. A robust conceptual comparison, such as ICA, therefore helps reveal where outputs converge or diverge in meaning (see Appendix 3 for a more detailed example). \\ 

\subsubsection{Step 4. Quantify the Inductive Conceptual Rating (ICR)}
The final step in calculating the ICR is to quantify the results. Researchers should evaluate the conceptual presence or absence of concepts in the human baseline and LLM-generated outputs using the accuracy formula:

\[
\text{Accuracy} = \frac{\text{True Positives} + \text{True Negatives}}{\text{True Positives} + \text{False Positives} + \text{True Negatives} + \text{False Negatives}}
\]

This calculation produces a quantifiable interpretive rating of meaning accuracy (i.e., how closely GenAI outputs reflect RTA outputs) and truthfulness (i.e., whether GenAI outputs preserve contextual relationships and produce supported or misleading interpretations). After evaluation, an ICR score ranging from 0 to 1 can be assigned. Higher values indicate stronger overall alignment with the human baseline, fewer errors (both false positives and false negatives), and better preservation of relevant concepts. This metric captures gradations of alignment by using the RTA output as the reference against which GenAI outputs are compared and measured.\\

For instance, suppose a researcher is evaluating an LLM-generated summary of a short text about climate change. The human baseline (RTA) identifies the key concepts: greenhouse gases, global temperature rise, renewable energy, and policy interventions. The LLM-generated output mentions greenhouse gases, global temperature rise, and renewable energy, but omits policy interventions and incorrectly adds carbon taxes are universally adopted. Using ICR, the researcher would calculate:
\begin{itemize}
    \item True positives (TP): 3 concepts correctly captured (\emph{greenhouse gases, global temperature rise, renewable energy})
    \item False negatives (FN): 1 concept missing (\emph{policy interventions})
    \item False positives (FP): 1 concept incorrectly added (\emph{carbon taxes are universally adopted})
    \item True negatives (TN): Not applicable here; no true negatives identified
\end{itemize}
In this example, the resulting ICR score is 0.6, which indicates that 60\% of the concepts identified in the GenAI output align with those present in the human baseline, while 40\% represent unsupported, potentially misleading, or hallucinatory additions. Interpreted in this way, the ICR score provides a quantitative indicator of conceptual alignment and contextual truthfulness between human-generated and GenAI-generated interpretations. A more detailed illustration of this scoring process is provided in Appendix 4. \\

\subsection{Positioning and Practical Implications}

It is important to acknowledge that the ICR metric is not intended to replace automated metrics, but rather to complement them in contexts where evaluation requires sensitivity to meaning, interpretation, and use. This is particularly critical for large language model tasks such as summarization and thematic generation, as well as for high-risk applications where truthfulness, contextual accuracy, and semantic fidelity are essential. Unlike automated metrics that treat meaning as a stable output derived from fixed inputs, ICR explicitly attends to meaning as a dynamic relation shaped by context, interpretation, and analytic judgment. Consequently, the effective application of ICR requires qualitative competency. In particular, to successfully apply ICR, researchers should have qualitative evaluative expertise and be familiar with rigorous hermeneutical practices that make interpretive decisions transparent and accountable. By integrating human interpretive insight with systematic assessment, ICR offers epistemological value by revealing how meaning is constructed, negotiated, and transformed across contexts, while also providing practical benefits by generating actionable insights for model evaluation, responsible deployment, and informed use. \\

\section{Case-Study: Analyzing Unstructured Text Data using the ICR Metric}

Unsurprisingly, given the premise of this article, we argue that understanding how large language models (LLMs) analyze unstructured text requires both linguistic and semantic evaluation. While LLMs can generate coherent outputs, the alignment of these outputs with human interpretations of meaning remains uncertain, particularly for complex or nuanced concepts. By using an ICR metric, to compare human-generated outputs with LLM-outputs, in this example, thematic summaries, we assess not only surface-level textual similarity but also deeper conceptual fidelity, interpretive alignment, and context. The approach enables rigorous evaluation across multiple datasets, models, and prompt strategies, providing insights into the accuracy, consistency, truthfulness, and limitations in machine-generated sense-making.

\subsection{Methodology}

This study used a mixed-methods triangulation design to assess the accuracy of LLM-generated thematic summaries \cite{Creswell2017}. Two data types were analyzed: qualitative open-text comments and quantitative ratings of those comments. The quantitative analysis followed a post hoc design using previously collected survey data, measuring textual similarity between original reference text and their corresponding LLM-generated summaries, followed by descriptive statistics and correlation tests. The qualitative analysis applied a Reflective Thematic Analysis (RTA) to generate outputs that reflect the reference datasets. Next, an Inductive Content Analysis (ICA) was used to compare the concepts and meanings in the original reference text with the RTA-generated outputs and the LLM-generate outputs. The RTA analyses were conducted first, before the LLM-generated outputs. Once the RTA and LLM-outputs were generated independently of each other, we integrated an ICR to evaluate surface-level (linguistic) similarity and deeper-level (semantic) meaning \cite{FitkovNorris2023, Guetterman2018}.

\subsubsection{Dataset}
Five datasets, ranging from N=50 to N=800 (see Table 2), were drawn from open-text survey responses pertaining to a question about perceptions of work. Each comment ranged from 1 to 300 words. Preprocessing included removal of personally identifiable information and exclusion of low-quality responses (e.g., “none,” “N/A”). To ensure consistency, both human analysts and LLMs were required to produce exactly three themes per dataset, each with a name (<5-words) and a 3 to 4 sentence summary. LLMs were prompted to output this standardized format in JSON to enable direct comparison with human-generated summaries. A synthetic example of the format is shown in Appendix 2. 

\begin{table}[ht!]
\centering
\caption{Characteristics of the Datasets Used in the Case Study: Response Counts, Corpus Size (word and character counts), and Prevalent Topics in Dataset}
\label{tab:dataset_details}
\small
\begin{tabularx}{\textwidth}{cccc>{\raggedright\arraybackslash}X}
%\hline
\toprule
\textbf{Dataset \#} & \textbf{Responses} & \textbf{Word Count} & \textbf{Character Count} & \textbf{Most Frequent Topics} \\ %\hline
\toprule
1 & 50 & 1,801 & 8,496 & Organization, leadership, strategy \\ %\hline
2 & 100 & 5,339 & 26,897 & Policy, organization, leadership \\ %\hline
3 & 200 & 7,467 & 36,221 & Culture, organization, strategy \\ %\hline
4 & 400 & 8,299 & 40,004 & Resources, organization, tasks \\ %\hline
5 & 800 & 49,982 & 238,389 & Organization, culture, leadership \\ %\hline
\bottomrule
\end{tabularx}
\end{table}

\subsubsection{Establishing Ground-truth}
To establish a ground truth, researchers conducted a Reflective Thematic Analysis (RTA) on each dataset using Braun and Clarke’s (2006) six-step method \cite{BraunClarke2006}. This served as a baseline to assess the accuracy, consistency, and potential biases in LLM-generated summaries. Multiple raters were involved to enhance reliability, and comparisons between human and LLM outputs were made using inductive conceptual content analysis (ICA) following Zhang and Wildemuth’s (2009) process \cite{ZhangWildemuth2009}. Concepts were coded using short phrases (<5 words) and categorized inductively. Consistency was evaluated through iterative coding, with discrepancies addressed by refining categories. The presence or absence of each concept was documented for comparison across outputs. \\

\subsubsection{LLM Selection and Implementation}
Examining Sonnet 3.5 and Nova Pro is valuable because they differ in architectural maturity, training emphasis, and representational capacity, which directly shapes how meaning is constructed in generated text \cite{Jacas2025Architecture}. Sonnet 3.5 emphasizes instruction-following and structured reasoning, making it useful for examining how prompts are translated into explicit thematic organization, while Nova Pro is comparatively less performant on standard benchmarks and often exhibits weaker abstraction and contextual integration \cite{Bedemariam2025Potential}. Studying these models together allows the ICR framework to capture gradations of meaning alignment across systems and demonstrates that high linguistic fluency does not necessarily equate to semantic accuracy, reinforcing the need for human-centered interpretive evaluation. \\

\subsubsection{Prompt Engineering}
Dozens of prompts were tested using various strategies (e.g., zero-shot, few-shot, chain-of-thought). We refined prompts iteratively based on output alignment with study objectives \cite{He2024Prompt}. Due to model-specific behavior, final prompts were tailored per model while maintaining a consistent structure (e.g., generate 3 themes, 2–4 sentence summaries). All models were run with standardized parameters: Temperature = 0, top-k = 0.25, top-p = 0.99. Importantly, each prompt and model was run ten times. In total, fifty outputs were generated for each model, with a total of 100 outputs \cite{He2024Prompt}. The most frequently recurring output used as the representative output for each model and dataset. \\

\subsubsection{Evaluation Metrics}

We used several key metrics across two dimensions: linguistic similarity and semantic accuracy. For linguistic metrics, we used cosine similarity and BERTScore. For semantic similarity, we used the ICR metric to measure agreement in concept presence between RTA and GenAI outputs. \\

\section{Results}

The results showed variability across models and datasets in terms of similarity, accuracy, and reliability. This section provides a high level overview of our findings while the following figures illustrate the results visually:
\begin{itemize}

\item \textbf{Dataset N = 50:} For the smallest dataset, all outputs show identical Recall, Precision, and F1 scores, indicating similar surface-level coverage. However, semantic accuracy diverges with the human-RTA achieving perfect ICR (1.00), while both Sonnet 3.5 and Nova Pro scored substantially lower (0.69). This suggests that although LLM-generated thematic summaries appear linguistically adequate, in this instance, the models missed or distorted core meanings, particularly with a small dataset.

\item \textbf{Dataset N = 100:} At N = 100, Sonnet 3.5 achieved high linguistic performance (Cosine = 0.89, F1 = 0.91), outperforming the human RTA across lexical metrics. Despite this, its ICR dropped to 0.35, indicating semantic misalignment. Nova Pro showed slightly lower linguistic scores but a higher ICR score (0.48). The human RTA maintained a strong balance between linguistic alignment and semantic accuracy (ICR = 0.86).

\item \textbf{Dataset N = 200:} With moderate data volume, linguistic metrics were relatively high across all outputs. Sonnet 3.5 and Nova Pro demonstrated strong Recall and F1, but semantic accuracy remained inconsistent with Nova Pro achieving substantially higher ICR rating (0.75), while Sonnet 3.5 had a lower score (0.39). The RTA demonstrated near-complete semantic coverage (ICR = 0.96).

\item \textbf{Dataset N = 400:} At N = 400, both LLMs achieved high linguistic similarity, particularly Sonnet 3.5 (F1 = 0.90). However, ICR scores were modest (0.47 for Sonnet; 0.53 for Nova), which indicates a gap in contextual meaning. The human RTA had a slight drop in Recall but maintained high semantic fidelity (ICR = 0.94).

\item \textbf{Dataset N = 800:} For the largest dataset, both models demonstrated their highest overall performance. Linguistic similarity was high across the board, and ICR scores improved (0.65 for Sonnet 3.5; 0.76 for Nova Pro). Nevertheless, the human RTA still achieved higher semantic accuracy (ICR = 0.93). These results suggest that increased data volume enhances LLM semantic alignment but does not fully close the gap with human interpretive analysis.

\end{itemize}

\begin{figure}
    \centering
    \caption{Evaluation Metric Results Visuals Comparing Human RTA Results with Sonnet 3.5 and Nova Pro}
    \includegraphics[width=\textwidth]{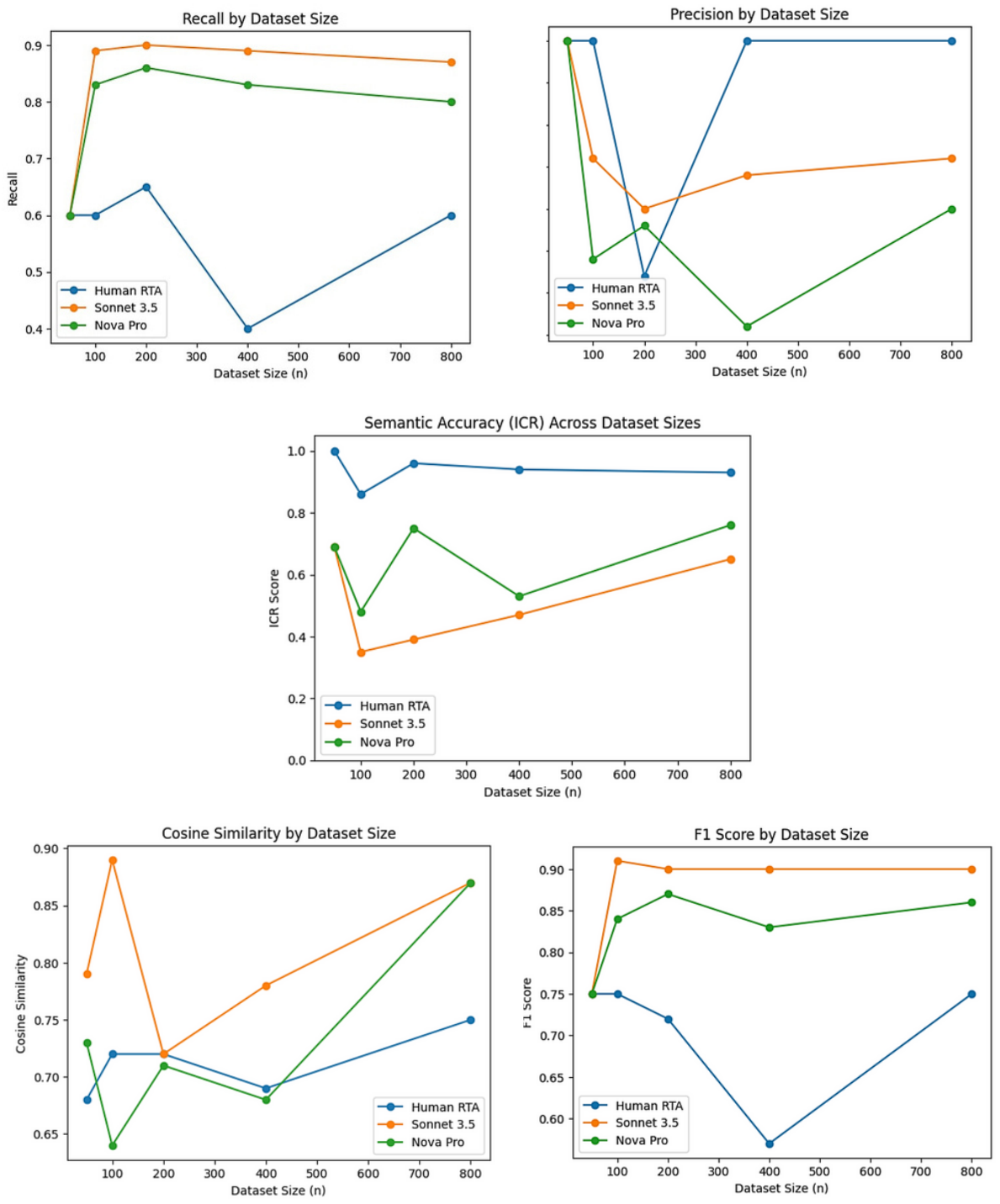} % Use your PDF file
    \label{fig:results_metrics}
\end{figure}

\section{Discussion}

This case-study illustrates a comparison between GenAI outputs and human-generated outputs of unstructured data meaning, using the ICR metric to quantify semantic fidelity. Across all dataset sizes, LLMs demonstrated high surface-level performance, which is shown by cosine similarity and F1 scores, but the models consistently had lower scores with semantic alignment. This divergence highlights a critical finding that LLMs can be effective at content matching, with respect to replicating lexical patterns and topical overlap, but these types of models may exhibit limited capacity for meaning matching, which requires relational reasoning, contextual grounding, and recognition of recurring and fluid conceptual structures. These results contribute to ongoing research about the interpretive capabilities of AI, reinforcing the semiotic distinction between signifiers (words) and signifieds (meaning). \\

We identified dataset size as a factor that seemed to influence semantic stability. While surface-level metrics remained consistent, ICR scores fluctuated substantially with smaller datasets and stabilized only once datasets exceeded approximately ~200 responses. Even with the largest dataset (N = 800), neither LLM fully reached human levels of interpretive coherence, suggesting that increased data volume improves but might not entirely guarantee semantic fidelity. Human researchers, in contrast, maintained high ICR scores regardless of dataset size, illustrating the robustness of human interpretive reasoning in extracting meaning from sparse data. This finding highlights both the strengths and limitations of AI-driven qualitative synthesis and reinforces the importance of incorporating human judgment. \\

Model-level analysis also revealed a nuances. While research suggests more advanced LLMs can achieve higher results across benchmarks \cite{Lunardi2025}, our study supports that models can achieve high linguistic metric results, but the models' demonstrated substantial variability when attempting to synthesize and generate accurate meaning from complex, multifaceted human experiences. These findings suggest that model newness or architectural complexity may not necessarily produce better results. This finding challenges the assumption that newer or larger models, as a metric like ICR can capture misalignment in LLM-generated outputs. \\

Lastly, the results reveal epistemological importance. Unlike traditional analytical tools, which primarily observe or measure a phenomena, the probabilistic nature of LLMs means these tools transform sign systems without reflexivity, historical situatedness, or awareness of cultural and contextual factors central to human understanding, which are important elements of traditional qualitative research and meaning-making \cite{CreswellPoth2016, Creswell2017, BraunClarke2006, ZhangWildemuth2009, SmithFlowersLarkin2009, MilesHubermanSaldaña2014}. The empirical ICR gap (researcher: 0.93 vs. LLMs: 0.35-0.76) quantifies this deficit and supports the view that GenAI outputs simulate rather than generate meaning \cite{Hicks2024}. This distinction is critical for researchers who seek to employ generative AI in interpretive work, particularly where high surface or content similarity does not imply deep comprehension of meaning. \\

\subsection{Implications}
Given the findings of the case-study, we identified several implications for theory, methodology, and practical use:

\begin{itemize}

\item \textbf{Lexical vs. semantic alignment:} This study found two foundation LLMs excelled at lexical, content matching but struggled with meaning matching. This suggests a friction between linguistic relatedness and interpretive understanding. The data reveals that automated metrics might get lexical alignment right but not necessarily the depth of meaning. Metrics such as ICR can help with assessing semantic fidelity beyond surface-level similarity.

\item \textbf{Dataset size and interpretive stability:} Semantic accuracy improved with larger datasets, yet even higher data volumes did not guarantee convergence with human interpretation. This study suggests researchers remain necessary to quantify meaning representation with GenAI outputs, especially in small-scale or conceptually nuanced unstructured text thematic syntheses.

\item \textbf{Epistemological considerations:} LLMs simulate meaning without reflexivity or contextual awareness. The gap between human and AI semantic performance highlights the importance of positioning GenAI outputs as tools for pattern detection and preliminary synthesis rather than as sources of understanding.

\item \textbf{Inductive evaluative approach:} Integrating LLMs with human baseline comparisons allows for efficient detection of patterns while preserving interpretive rigor. ICR provides a transparent framework to assess GenAI outputs and offers a validation process to evaluate meaning. It also requires experienced human researchers who are proficient in multiple analytical approaches to be conducted effectively. 

\end{itemize}

Overall, our findings indicate that the two LLMs examined in this study can generate surface-level lexical matching outputs, but the models struggled with deeper, context-dependent semantic meaning. With that said, metrics like ICR indicate value in measuring semantic fidelity to determine the extent that an LLM is truthful in its outputs or not. \\

\subsubsection{Limitations}

We acknowledge several limitations associated with the case-study. First, the case study focused on open-text survey responses to a single question about perceptions of work; the results may not generalize to other domains or data types. Second, the establishment of ground truth through Reflective Thematic Analysis (RTA), while methodologically rigorous and supported by multiple raters, reflects a situated human interpretive framework rather than a universal representation of meaning, which is reflective of the human condition in interpreting signs, signifieds, and signifiers. Third, the analysis examined only two language models using tailored prompts and fixed parameters, and the observed patterns of semantic misalignment may be due to model or prompt-specific instructions and may not extend to other models or configurations. Fourth, the ICR metric operationalizes semantic alignment through binary concept presence, which captures conceptual coverage but does not account for degrees of overlap, relational structure, or differences in interpretive emphasis; there are opportunities to expand this metric further. These limitations suggest that while the ICR framework offers a promising approach for evaluating semantic fidelity in GenAI outputs that focus on thematic summarization or more broadly conceptual analysis, further validation across domains, models, and contexts is recommended for future research. \\

While the use-case in this study illustrated unstructured survey text-based data, ICR’s transferability to other research domains and fields relies on the type of content and extent of human expertise needed. Conceptually, the method is domain-agnostic, as it has been designed to compare GenAI-outputs with a human-generated interpretive baseline. This allows for a range of settings where themes, concepts, or semantic relationships are present and can be meaningfully coded. However, there are specific instances where ICR might require additional considerations. For one, domain-specific language or settings may require specific adjustments. Legal texts, for instance, might contain subtle differences in phrasing that may change the nature of rights or policies, and an RTA baseline must include fine-grain distinctions. Alternatively, clinical notes will likely contain shorthand, acronyms, and terminology that may require coders to be experts in the field to avoid misinterpretation.Aside from field-considerations, ICR largely depends on human-coded baselines, which requires that coders contain domain and qualitative expertise to be able to identify relevant concepts accurately. Since LLMs can behave differently across various domains, foundational model architecture might misrepresent legal jargon or medical terms or cultural dialect, and an ICR can be ideal in quantifying these divergent ruptures of meaning, but the baseline must reflect domain-specific meaning to ensure more valid alignment. Overall, the ICR metric is highly adaptable across different domains, but its success depends on careful human baseline creation, domain-aware coding practices, and can be enhanced with inter-rater reliability. \\

\section{Conclusion}

This study highlights GenAI variability in surfacing meaning and the ways an ICR metric allows researchers to avoid measuring stimulated meaning, and, instead use actualized meaning to determine output accuracy. While the LLMs examined in this study demonstrate impressive proficiency in replicating surface-level lexical patterns, these models can fall short in achieving semantic depth and relational coherence that characterize human interpretation; this distinction highlights a fundamental tension between statistical fluency and interpretive understanding, revealing that the replication of language is not synonymous with grammar or word-matching. \\

From a theoretical perspective, these findings reinforce the semiotic distinction between signifiers and signifieds, and they echo hermeneutic principles emphasizing contextuality, reflexivity, and the situated nature of meaning. GenAI, by probabilistically transforming sign systems, simulates interpretive processes but is inhibited by the historical, cultural, or experiential context necessary for genuine meaning understanding. The empirical gap in ICR scores between humans and LLMs in this study quantifies the epistemic limitations and offers a lens in which researchers can assess the interpretive reliability of AI analysis. \\

We would be remiss if we did not recognize the multidimensionality challenge of meaning. As Picasso (1966) observed, “If there were only one truth, you couldn’t paint a hundred canvases on the same themes.” Picasso hits on the inherently messy nature of meaning; it can be plural, layered, nuanced, and open to multiple expressions. In this light, LLMs, in producing varied but imperfect approximations of human understanding, may offer a single or perhaps even a few different representations of meaning. Albeit, those representations may be inherently wrong, but they could also be accurate; without a baseline to compare, it is difficult to fully know. To this end, we argue LLMs should not serve as the composite truth-teller or meaning maker. Ultimately, we advocate for an inductive epistemology, where researchers provide interpretation, contextual grounding, and ethical reflection that are necessary for systematic meaningful analysis. By using metrics like ICR, researchers can draw clearer boundaries between identifying linguistic simulation and deeper conceptual understanding. In doing so, researchers can ensure that GenAI outputs enhance, rather than simulate, or worse, obscure the richness of human meaning-making. \\

%Bibliography
\bibliographystyle{unsrt}  
\bibliography{references}  

% Appendices
\appendix

\section{Appendix 1: Additional Key Concepts}

\subsection{Semiotics and Meaning}
The field of semiotics  posits that meaning does not reside in words themselves but emerges through relational systems of signs. Saussure, who in 1916 described a linguistic sign to be composed of the signifier and the signified, argued that meaning does not emerge from a direct correspondence between a word and a concept; instead, meaning emerges from the position of a sign within a broader system of differences. Meaning, therefore, is inherently relational and context-dependent rather than fixed or intrinsic. He posits that humans have an incredible ability to fluidly recognize and interpret signs, given a variety of conditions. In later years, 'sembiotic theorists expanded Saussure’s early theories to include the instability and multidimensionalness' of meaning. For instance, Barthes (1977) argued that signs function within cultural or ideological systems \cite{Barthes1977}. Within these particular systems, meaning is constantly reinterpreted and reshaped. From this perspective, theorists suggest that a sign is never neutral in meaning, but rather meaning is always influenced by social, historical, and/or cultural contexts in which the signs reside in and are interpreted from. From this perspective, the same sign may mean something in one culture and yet the same sign may mean something entirely different in another culture. \\

In talking about the concept of culture, we must recognize micro, meso, and macro layers. For instance, even subtle cultural differences can influence sign meaning. In fact, a sign could mean something distinct within a micro-culture, such as a particular group of people, that function under a broader cultural or population context, such as home culture compared to national culture \cite{hofstede2001culture}. Take the word “family.” Within a micro-culture, such as a particular Hawaiian 'ohana, “family” can include or refer to not only parents and children but also extended relatives, close friends, and even community members who share mutual care and responsibility \cite{kanuha2005ohana}. Whereas, in the broader U.S. national culture, “family” is often interpreted more narrowly as the nuclear unit of parents and children. Globally, in other cultural contexts, “family” might extend to multiple generations living together, or carry legal and social connotations that differ entirely from U.S. usage. Signs across these cultural examples can thus vary in meaning and also contain overlap, depending on cultural context. Hence, signs, and more broadly meaning, are dependent on the conditions of the system they reside in, rather than a word itself. \\

\subsubsection{Polysemy}
The concept of polysemy is the study of the multidimensionality of a sign. In other words, polysemy is one signifier (word) that is associated with multiple signifieds (meaning), such as the word “nevermore”  in Poe’s The Raven \cite{Lyons1977, Cruse2004}. Words that contain multiple meanings, such as the contextual positioning of the word “nevermore” or the more popular polysemy word “run” do not contain a single, stable meaning; instead, their interpretation often depends on contextual cues, narrative positioning, and relational use. Run, for instance, has hundreds of meanings, including moving fast, flowing, making a regular journey, managing, a candidate in a political election, etc. When we consider this duplicity with words, semiotics become particularly important and informative in language and meaning. Afterall, humans interpret by decoding language fluidly, and recognizing the sign, signified, and signified allows us to engage in interpretive flexibility, which is a central part of how human language functions \cite{DeSaussure1916}. \\

\subsubsection{Computational Representations of Meaning}
In computational linguistics and natural language processing (NLP), meaning is modeled through statistical representations of language. Word embeddings are used to encode words as vectors in high-dimensional vector spaces that allow for semantic clustering by grouping words in patterns of co-occurrence and distributional similarity \cite{Li2024, Dvorackova2025, Apidianaki2023, Izzidien2022}. Traditional and later static embedding models include Bag of Words, TF-IDF, Word2Vec, and GloVe, and these models assign a single, fixed vector to each word type, regardless of context \cite{Alkaabi2025}. Consequently, these models operate at the level of signifiers (words) rather than signifieds (meaning), as they can capture a word, but not necessarily its situated meaning \cite{ZhangPeng2024}; this means traditional NLP models output language where meaning is inferred indirectly through frequency and proximity relationships rather than through conceptual or contextual understanding, which has resulted in challenges with polysemy and contextual meaning \cite{Bhargavi2025}. \\

\subsubsection{Contextual Embeddings}
Recent advances in embedding models have sought to address traditional embedding limitations through contextualized embeddings. In models such as BERT, RoBERTa, ALBERT, and DistilBERT, each input token passes through multiple self-attention and feedforward layers, where a token is iteratively updated to reflect its contextual meaning. For example, the word “run” is interpreted differently when used in the context of a “model run” versus a “marathon run,” even though the “run” token embedding is identical; the model’s transformations disambiguate meaning according to the context of “model” and “marathon” \cite{Bhargavi2025, Sengar2025}. This process allows embeddings to capture fine-grained semantic distinctions that static embeddings cannot, improving performance in tasks like topic modeling, summarization, and sentiment analysis \cite{Viegas2025, GangundiSridhar2025, PaneruThapa2025}. From a semiotic perspective, the contextual embedding process reconfigures the relationship between signifier and signified. While promising, gaps are still present in meaning representations with contextual embedding models. This is not entirely surprising because human meaning identification emerges through interpretive engagement with signs in context, whereas GenAI approximates meaning via statistical relationships in latent vector spaces, where similarity, distance, and direction encode semantic relatedness \cite{DeSaussure1916, Martin2024, Apidianaki2023}. Despite their advantages, contextual embeddings are statistical approximations, grounded in mathematical patterns rather than fluid, embodied, culturally situated and interpretive understanding. \\

\subsubsection{Transformer Layers and Meaning}
In large language models, transformer layers operationalize contextual embeddings using self-attention mechanisms to weigh the relevance of each token relative to others, while feedforward networks project these weighted representations into higher-dimensional spaces that encode increasingly abstract semantic features \cite{Li2024}. Through this iterative transformation, polysemous words gain context-specific vector representations, allowing the model to distinguish multiple signifieds (meanings) of the same signifier (word). However, these processes are still statistical approximations, since they capture semantic coherence probabilistically without actualized interpretive understanding, making GenAI outputs sensitive to prompt phrasing and probabilistic sampling \cite{Requeima2024, Liu2025}. From a semiotic perspective, transformer layers and contextual embeddings enable GenAI models to simulate the relational dynamics between signifier and signified, but the models do not interpret meaning in the traditional human linguistic or hermeneutic sense \cite{DeSaussure1916, Apidianaki2023, Martin2024}; because models reproduce how words tend to be connected to meanings in language, but the models do not interpret meaning, since the model does not inherently understand, interpret, or reflect the way humans do through a bricolage of lived experience and intention. \\

\subsubsection{Conceptual Variability and Hermeneutics}
While contextual embeddings and transformer layers allow language models to disambiguate polysemous words at the sentence level, challenges remain across longer sequences, where multiple interpretations of interacting words coexist as probabilistic possibilities in latent space \cite{Devlin2019, HaberPoesio2024}. For example, consider the concept of “work-life harmony.” If fifty individuals provided their personal experiences of the concept, and a researcher prompted a language model to define work-life harmony given the input data of the 50 individuals, the model would synthesize these inputs by negotiating statistical patterns encoded in embeddings and transformer layers, albeit with some contextual constraints from the prompt. The resulting output might emphasize dominant themes, but there is a risk of underrepresentation of less frequent perspectives, or subtle variations in phrasing or context that could yield different interpretations from a placed-based or contextual positionality from an individual embedded within a particular culture or group. Although transformer-based models produce coherent text, the models do not “understand” meaning in a human sense and remain probabilistic, surfacing approximations of semantic relationships. GenAI outputs ultimately are an advanced guessing-game of what a text’s meaning might be, given certain conditions. \\

Alternatively, hermeneutics is a critical framework for understanding and identifying meaning and meaning variability in human experiences. Gadamer (1975) emphasizes that interpretation is historically and culturally situated, shaped by the interpreter’s prior knowledge, cultural horizons, and the social context in which meaning is produced \cite{Gadamer1975}. Alternatively, Ricoeur (1970) argues that there are different types of hermeneutics, such as a hermeneutic of suspicion, which critically interrogates underlying structures, assumptions, and power dynamics, and a hermeneutics of faith, which seeks to identify coherence, meaning, and narrative continuity within texts \cite{Josselson2004}. Applied to a dataset of individual experiences, Gadamer and Ricoeur's perspectives of hermeneutics encourages researchers to engage in a layered approach, where one can explore patterns, recurring themes, and implicit assumptions (suspicion), while also attending to the lived coherence and subjective sense-making expressed by individuals (faith). \\ 

From a hermeneutical perspective, like a semiotic lens, meaning is never fixed; instead, meaning emerges through a dialogical process of interpretation, as the researcher actively negotiates between the signifiers (words), the signified (meaning), and the broader social, cultural, and temporal context within a text. Applied to the example of 50 individuals, hermeneutical practices can highlight what work-life harmony means from diverse human experiences; this approach allows researchers to synthesize multiple perspectives of “work-life harmony,” revealing both shared elements and unique variations, while situating each individual response within a dynamic, relational framework of meaning. It also forces researchers to examine “worklife harmony” from a nomothetic and an idiographic perspective, moving between the whole and the individual, the general and the specific. Hermeneutics forces researchers to recognize potential friction or tension in experiences, and identify where experiences overlap and differ to avoid over-indexing on a dominant view or non-dominant view. From this lens, hermeneutics complements semiotic theory by foregrounding the interpretive work required to translate raw textual data into coherent, contextually grounded understanding, rather than treating meaning as a static property inherent in data. \\

\subsubsection{Challenges with GenAI Outputs and Meaning Representation}
Despite the advantages of contextual embeddings and transformer architectures, GenAI outputs face fundamental limitations in representing meaning, particularly when dealing with complex, multi-faceted human experiences. From a semiotic and hermeneutic perspective, GenAI outputs reveal a tension between signifiers (words) and signifieds (meaning). For instance, embeddings can simulate semantic relationships \cite{Nesbitt2024} but lack reflexive interpretation \cite{Shan2025}; since LLMs mimic language through highly complex mathematical systems and processes but do not aim to analyze data grounded in identifying truth or the reality of human experiences, these models may produce biased or misleading interpretations without recognizing them as such \cite{Resnik2025}. This means evaluating GenAI requires frameworks that integrate technical understanding with interpretive methods, such as reflective thematic analysis and inductive content analysis, to trace how meaning emerges, shifts, or diverges from human experience. Treating GenAI as a semiotic infrastructure highlights that outputs simulate rather than understand meaning \cite{Hicks2024}. Ultimately, in viewing GenAI this way highlights that outputs are simulations of meaning, structured by the model’s architecture and training data, and not true understanding. Therefore, human interpretations are critical to assess meaning fidelity, coherence, and truthfulness.

\section{Appendix 2: Example of GenAI Thematic Outputs in JSON}
theme1: 
"name": "Adaptive Organizational Learning", \\
"summary": "Organizations are increasingly prioritizing continuous learning as a core capability. By encouraging experimentation, reflection, and knowledge sharing, teams are better equipped to respond to uncertainty and change. This adaptive approach allows organizations to evolve alongside emerging challenges while cultivating resilience and long-term innovation." \\

theme2:
"name": "Human-Centered Technology Integration", \\
"summary": "As advanced technologies become embedded in everyday work, organizations are re-centering human values in system design and implementation. Emphasizing usability, transparency, and ethical considerations ensures that technology supports rather than replaces human judgment. This integration strengthens trust, efficiency, and meaningful engagement between people and tools." \\

theme3:
"name": "Distributed Leadership Practices", \\
"summary": "Leadership is shifting from hierarchical control toward distributed and collaborative models. Teams are empowered to make decisions, share responsibility, and contribute diverse perspectives. This redistribution of leadership fosters accountability, accelerates problem-solving, and supports more inclusive organizational cultures." \\

\section{Appendix 3: RTA Procedural Example}

The study followed a structured Reflective Thematic Analysis (RTA) process. The steps are outlined below:

\begin{enumerate}

\item \textbf{Data Familiarization}  
Researchers first read all 50 participant responses multiple times to immerse themselves in the data. Sample excerpts include:
\begin{itemize}
    \item “I feel balanced when I can manage my deadlines without sacrificing family time.”
    \item “Sometimes work demands spill over into my personal life, which makes me anxious.”
    \item “Having flexible hours allows me to pursue hobbies and maintain relationships.”
    \item “I constantly struggle with guilt when prioritizing my career over personal time.”
\end{itemize}
This stage allowed the research team to note initial impressions, recurring concepts (e.g., \emph{balance}, \emph{flexibility}, \emph{guilt}), and variations in experiences.

\item \textbf{Initial Coding}  
Researchers applied open coding to the data, assigning descriptive labels to meaningful units. Example codes include:
\begin{itemize}
    \item \emph{Boundary management}: describing the ability to separate work and personal life
    \item \emph{Flexibility}: references to adjustable schedules or remote work options
    \item \emph{Emotional tension}: feelings of guilt, stress, or anxiety related to competing priorities
    \item \emph{Personal fulfillment}: engagement in hobbies, family, or self-care
\end{itemize}
Multiple researchers coded independently to capture diverse interpretations.

\item \textbf{Theme Identification}  
Codes were grouped into broader conceptual themes:
\begin{itemize}
    \item \emph{Work-Life Balance}: managing competing demands of work and personal life
    \item \emph{Autonomy and Flexibility}: control over schedule, ability to work remotely, or choose tasks
    \item \emph{Emotional Experience}: stress, guilt, satisfaction, or relief associated with harmonizing roles
    \item \emph{Meaning and Purpose}: sense of fulfillment derived from achieving harmony in daily routines
\end{itemize}

\item \textbf{Thematic Review}  
Themes were reviewed for coherence and overlap. Researchers examined whether each theme adequately captured the coded data and maintained conceptual clarity. For example, some codes related to stress were initially grouped under “Emotional Experience” but were subdivided into work-induced stress and life-induced stress for nuance.

\item \textbf{Defining Themes}  
Each theme was precisely defined and contextualized:
\begin{itemize}
    \item \emph{Work-Life Balance}: Practices and strategies participants use to allocate time and energy between professional and personal responsibilities
    \item \emph{Autonomy and Flexibility}: Extent to which participants can make choices about work timing, location, and task prioritization
    \item \emph{Emotional Experience}: Positive and negative affective responses arising from attempts to achieve harmony
    \item \emph{Meaning and Purpose}: Subjective interpretations of fulfillment, identity, and satisfaction from integrating work and life roles
\end{itemize}

\item \textbf{Generating Insights}  
The final RTA output highlighted relationships among themes. For example, participants with higher \emph{Autonomy and Flexibility} reported lower \emph{Emotional Tension} and higher \emph{Meaning and Purpose}, suggesting that structural supports directly influence experiential outcomes.

\end{enumerate}

\subsubsection*{Inter-Rater Reliability}  
Cohen’s kappa ($\kappa = 0.87$) and Fleiss’ kappa ($\kappa = 0.85$) indicated high agreement among researchers, confirming consistent thematic interpretation.

\subsubsection*{Interpretive Baseline}  
The resulting thematic structure, detailing conceptual sequencing, relationships, and contextual nuance, constitutes the human-interpreted baseline. This baseline provides the benchmark for subsequent ICR evaluation of GenAI-output summaries or analyses of participants’ experiences with work-life harmony.

\section{Appendix 4: ICA Output Example}
\subsection{Inductive Content Analysis of GenAI Outputs}

The analysis followed a structured Inductive Content Analysis (ICA) procedure on the GenAI outputs. Steps are outlined below:

\begin{enumerate}

\item \textbf{Data Preparation}  
GenAI outputs were collected in response to the prompt: 
\emph{``Summarize the experiences of 50 participants discussing their work-life harmony.''} Sample outputs included:

\begin{itemize}
    \item “Participants prioritize career success, often at the expense of personal time, though some report flexible schedules aiding balance.”
    \item “Many participants experience stress and guilt, but hobbies and family activities help mitigate work pressures.”
    \item “Work-life harmony is achieved when individuals can manage deadlines while also dedicating time to self-care and relationships.”
\end{itemize}

The data were cleaned for consistency, segmented into meaningful units (sentences or phrases), and prepared for coding.

\item \textbf{Unit of Analysis}  
Each sentence or meaningful phrase was treated as a unit of analysis, enabling the researchers to capture distinct concepts or semantic contributions from the AI output.

\item \textbf{Developing Categories / Coding Scheme}  
Using inductive content analysis via Zhang et al. (2009), emergent categories were identified based on recurring patterns, without imposing RTA-derived themes. Example emergent categories included:

\begin{itemize}
    \item \emph{Career Emphasis}: mentions of work, deadlines, productivity
    \item \emph{Personal Well-being}: references to hobbies, family, relaxation, mental health
    \item \emph{Flexibility}: mentions of schedule control, remote work, adjustable hours
    \item \emph{Emotional Tension}: stress, anxiety, guilt associated with balancing roles
    \item \emph{Integrated Meaning}: statements about achieving harmony or fulfillment
\end{itemize}

\item \textbf{Testing and Coding the Dataset}  
Researchers iteratively coded several GenAI output samples to refine categories, resolve ambiguities, and ensure consistent interpretation. Once finalized, the full dataset of GenAI outputs was coded according to the emergent categories.

\item \textbf{Assessing Consistency}  
Inter-coder agreement was checked (Cohen’s $\kappa = 0.81$), confirming reliable coding of GenAI outputs.

\item \textbf{Comparison with RTA Baseline}  

The emergent ICA categories were compared to the RTA-derived human baseline themes. Table~\ref{tab:ica_vs_rta} summarizes the comparison:

\begin{table}[ht!]
\centering
\caption{Comparison of RTA Themes (Human), ICA (LLM) Categories and Analytical Assessment (Observations) of Semantic Alignment, Distortion, and Omission}
\label{tab:ica_vs_rta}
\begin{tabularx}{\textwidth}{>{\raggedright\arraybackslash}X>{\raggedright\arraybackslash}X>{\raggedright\arraybackslash}X}
%\hline
\toprule
\textbf{RTA Theme} & \textbf{ICA Category Representation} & \textbf{Observations} \\ %\hline
\toprule
Flexible Scheduling & Flexibility & GenAI outputs captured flexibility but emphasized remote work more than participant quotes indicated. \\ %\hline
Personal Well-being & Personal Well-being & GenAI outputs underrepresented nuanced emotional experiences, e.g., guilt and anxiety, in favor of generic ``self-care.'' \\ %\hline
Career-Personal Integration & Integrated Meaning & GenAI outputs produced overgeneralized statements about achieving harmony, sometimes combining career and personal life without reflecting participants’ tension. \\ %\hline
Emotional Experience & Emotional Tension & GenAI outputs acknowledged stress but often minimized the relational context present in participant responses. \\ %\hline
\bottomrule
\end{tabularx}
\end{table}

This comparison highlighted alignment, distortion, omission, and amplification. For example, GenAI outputs aligned with RTA in capturing broad themes (Flexibility, Career Emphasis) but underrepresented nuanced emotional experiences (guilt, anxiety) and overgeneralized integrative statements.

\item \textbf{Calculating the ICR Score}

The GenAI outputs were evaluated by comparing the conceptual content of the generated interpretations with the themes identified through Reflexive Thematic Analysis (RTA). Conceptual alignment was assessed using the following classification categories:

\begin{itemize}
    \item \emph{True Positives (TP)}: Concepts present in the RTA baseline that were accurately reflected in the GenAI output (e.g., \emph{Flexibility}, \emph{Career Emphasis}).
    \item \emph{False Positives (FP)}: Concepts introduced by the GenAI output that were not present in the RTA baseline (e.g., ``all participants enjoy career success'').
    \item \emph{False Negatives (FN)}: Concepts identified in the RTA baseline but omitted from the GenAI output (e.g., nuanced emotional tension or feelings of guilt).
    \item \emph{True Negatives (TN)}: The absence of unrelated or irrelevant conceptual content in the GenAI output.
\end{itemize}

Using these classifications, an Inductive Conceptual Rating (ICR) score was calculated using an accuracy-based metric. For example, in the current dataset, the GenAI output correctly captured 5 relevant concepts (TP = 5), introduced 1 unsupported concept (FP = 1), omitted 1 relevant concept (FN = 1), and had no additional irrelevant concepts (TN = 0). Using the accuracy formula:

\[
ICR (\text{Accuracy}) = \frac{TP + TN}{TP + FP + TN + FN} = \frac{5 + 0}{5 + 1 + 0 + 1} = \frac{5}{7} \approx 0.714
\]

This ICR score of approximately 0.714 reflects moderate-to-strong overall conceptual alignment between the GenAI interpretation and the human baseline; this indicates that most themes were captured accurately, while a smaller portion of identified concepts were either missed or incorrectly introduced. Higher ICR scores represent stronger correspondence with the human-generated analysis and fewer errors in conceptual interpretation.

\end{enumerate}

\end{document}